\documentclass[10pt]{article}
\usepackage[preprint]{tmlr}

\usepackage[utf8]{inputenc}
\usepackage[T1]{fontenc}
\usepackage{hyperref}
\usepackage{url}
\usepackage{booktabs}
\usepackage{multirow}
\usepackage{amsfonts}
\usepackage{nicefrac}
\usepackage{microtype}
\usepackage{xcolor}
\usepackage{amsmath}
\usepackage{amssymb}
\usepackage{subcaption}
\usepackage[most]{tcolorbox}
\tcbuselibrary{breakable}
\usepackage[capitalize,noabbrev]{cleveref}

\title{VEGAS: Human-Aligned Video Caption Evaluation via Gaze}

\author{\name Shenghui Chen \email shenghui.chen@utexas.edu \\
      \addr The University of Texas at Austin
      \AND
      \name Po-han Li \email pohanli@utexas.edu \\
      \addr The University of Texas at Austin
      \AND
      \name Ximeng Sun \email Ximeng.Sun@amd.com\\
      \addr AMD
      \AND
      \name Shijia Yang \email Shijia.Yang@amd.com\\
      \addr AMD
      \AND
      \name Emad Barsoum \email Emad.Barsoum@amd.com\\
      \addr AMD
      \AND
      \name Zicheng Liu \email Zicheng.Liu@amd.com\\
      \addr AMD
      \AND
      \name Sandeep Chinchali \email sandeepc@utexas.edu \\
      \addr The University of Texas at Austin
      \AND
      \name Ufuk Topcu \email utopcu@utexas.edu \\
      \addr The University of Texas at Austin}

\def\month{MM}  %
\def\year{YYYY} %
\def\openreview{\url{https://openreview.net/forum?id=XXXX}} %

\IfFileExists{../src/analysis/generated/human_gemini_pca_metrics_aea.tex}{%
    \input{../src/analysis/generated/human_gemini_pca_metrics_aea.tex}%
}{%
}

\definecolor{lightgray}{gray}{0.9}
\definecolor{datasetAcolor}{HTML}{A3C4F3} %
\definecolor{datasetBcolor}{HTML}{FBC4AB} %
\definecolor{datasetCcolor}{HTML}{C9E4DE} %

\usepackage{wrapfig}

\begin{document}

\maketitle

\begin{abstract}
Vision-language models excel at video captioning, yet typically generate descriptions that fail to capture individual viewers' attention. We propose VEGAS (\textbf{V}ideo caption \textbf{E}valuation via \textbf{GA}ze \textbf{S}core), a training-free metric that leverages test-time gaze to sample personalized, attention-aligned text. It is a cross-modal, information-theoretic metric that quantifies how well a candidate caption matches a viewer's focus.
To evaluate VEGAS, we curate a dataset of egocentric activities and instructional slides paired with synchronized gaze and reference annotations. We then select captions based on VEGAS via rejection sampling without model retraining.
Experiments show that VEGAS-selected captions align significantly better with human focus and improve downstream caption-to-video retrieval, demonstrating the practical utility of incorporating viewer attention during inference.

\end{abstract}

\section{Introduction}

In many applications, video captions serve not merely as descriptions but as interfaces to user intent.
Vision-language models (VLMs) generate semantically accurate captions, yet they are typically trained on crowd-sourced annotations that aggregate across diverse human interpretations. As a result, they often describe what is broadly visible while ignoring viewer-specific attention and subjective perception.
Consider an egocentric setting: two users observing the same kitchen scene may attend to different objects or actions, but a conventional VLM may produce a single globally correct caption that is pragmatically misaligned with what either user intends to retrieve, revisit, or act upon.
In caption-indexed retrieval, users often search for videos by referring to attended objects or actions rather than the full scene. When an index caption emphasizes unattended background content, the relevant video can become harder to retrieve.

\begin{figure}[t]
    \centering
    \includegraphics[width=\linewidth]{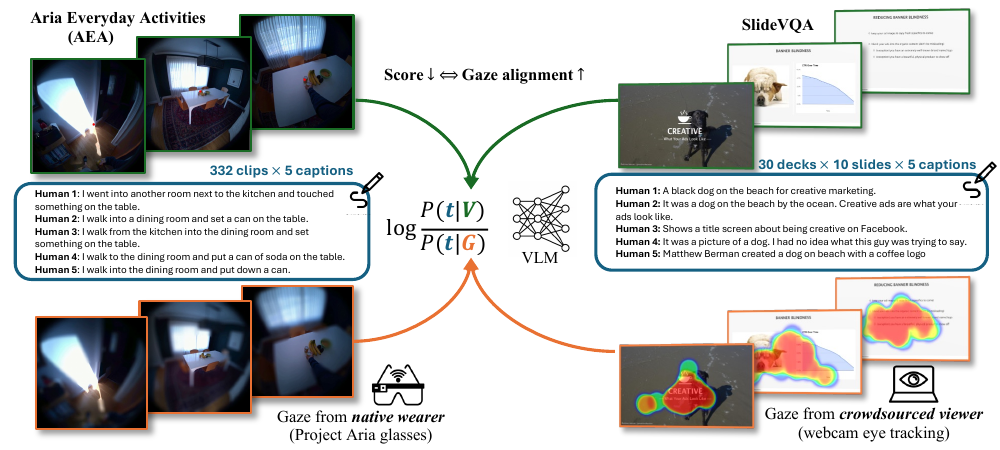}
     \caption{\textbf{Overview of our framework.} We curate a multimodal dataset pairing synchronized human gaze and annotated captions across egocentric video clips and instructional slides. VEGAS then leverages VLM's conditional probabilities to evaluate the alignment between a caption and a viewer’s gaze.}
    \label{fig:vegas_overview}
\end{figure}

Gaze provides a measurable, though imperfect, proxy for visual attention~\cite{just1976eye}. 
We use gaze at test time to select among plausible captions, favoring descriptions that better reflect the viewer's attended referents without retraining or directly modifying the captioning model.
Unlike prior work that uses human gaze only as a training-time supervision signal, we leverage gaze at test-time as a conditioning signal to generate personalized captions that reflect an individual's specific attention patterns. 
We introduce the \textbf{V}ideo caption \textbf{E}valuation via \textbf{GA}ze \textbf{S}core (VEGAS), a cross-modal, information-theoretic metric that quantifies how well a caption reflects a viewer's gaze patterns (\cref{fig:vegas_overview}). 
We then use VEGAS to select gaze-aligned captions via rejection sampling, without requiring model retraining.

To enable controlled evaluation across diverse video stimuli, we curate a multimodal dataset that aligns human gaze signals with visual content and human-written captions. The dataset spans two complementary domains: egocentric videos from the Aria Everyday Activities dataset (AEA)~\cite{lv2024aria} and instructional slide decks from SlideVQA~\cite{SlideVQA2023}. We choose these two domains because they represent two common gaze-driven scenarios: acting in the world and seeking information from dense visual media.
While several prior datasets include video-gaze or video-language annotations, datasets that jointly provide synchronized video, gaze, and captions are scarce; existing examples such as VAS~\cite{yu2017supervising} are relatively old and confined to movie clips.
Our dataset supports analysis of attention-driven semantic variation, evaluation of gaze-conditioned captioning, and benchmarking of models that aim to generate personalized, attention-aware video descriptions.

Experiments on our curated dataset provide the strongest support for VEGAS on AEA, where gaze helps resolve concrete object ambiguities.
Gaze-conditioned captioning yields a statistically significant increase in mean SBERT similarity (+0.0856), which translates to retrieval gains over conventional VLM-generated descriptions, improving mAP by +1.14\%, +2.48\%, and +2.46\% at ranks 1, 5, and 10, respectively.
The SBERT improvement is smaller and non-significant on SlideVQA (+0.0256).
This domain variance indicates that VEGAS is most effective when visual attention can be used to disambiguate concrete referents, but less diagnostic when the stimulus supports many valid conceptual summaries from the same visual.

\section{Related Works}

\paragraph{Captioning with Gaze.}
Video captioning has traditionally relied on CNN-RNN architectures, such as LRCN \cite{donahue2015long}, to model visual and temporal dynamics \cite{aafaq2019video}. 
To enhance localization and relevance, human gaze has been integrated as a supervisory signal in both image \cite{sugano2016seeing} and video captioning \cite{yu2017supervising}. 
Gaze-based supervision also increasingly supports object detection, action recognition, and skill assessment in egocentric and task-oriented contexts \cite{lall2025eyes,li2021eye,wu2025skillsight}, with recent efforts focusing on supervised fine-tuning with gaze annotations \cite{deng2024fine}.
Beyond training-time labels, recent studies also employ human gaze during inference for egocentric video understanding \cite{peng2025eye}, intent resolution \cite{madinei2026iris}, and difficulty-controlled text generation \cite{sauberli2026controlling}. 
While parallel research utilizes saliency prediction as a proxy for empirical gaze or evaluates the alignment between machine and human attention, our work treats gaze as a test-time conditioning signal to select captions aligned with individual attention patterns.

\paragraph{Video-Gaze-Language Datasets.}
The intersection of visual grounding and human attention has been facilitated by several key datasets that provide synchronized video, gaze, and annotated captions. Early efforts, such as VAS \cite{yu2017supervising} and Film Gaze \cite{breeden2017gaze}, integrated movie sequences with human-provided descriptions and gaze traces. Action recognition datasets, \textit{e.g.}, Actions in the Eye \cite{mathe2014actions} explores gaze patterns across task-specific viewing groups, and GTEA Gaze+ \cite{li2018eye} provides fine-grained action annotations for egocentric video. More recently, EgoExo4D \cite{grauman2024ego} offers a large-scale multimodal resource featuring egocentric perspectives paired with expert commentary and atomic action descriptions.
In contrast, our dataset provides synchronized video, human gaze, and free-form captions across more diverse domains, including egocentric and presentation slides, enabling a broader study of gaze-conditioned language understanding.

\paragraph{Evaluating Captions.}
Traditional metrics rely on gold-standard references, which are costly to obtain and not always reliable. 
Metrics, such as BLEU \cite{papineni2002bleu}, METEOR \cite{banerjee2005meteor}, and CIDEr \cite{vedantam2015cider}, quantify the lexical similarity between candidate and reference captions.
In contrast, metrics such as SPICE \cite{anderson2016spice} and S-BERT \cite{reimers2019sentence} assess deeper semantic similarities and structural alignment between captions.
Recently, cross-modal evaluation approaches propose information-theoretic evaluation by measuring the mutual information between video frames and captions \cite{chen2025vibevideototextinformationbottleneck, li2026visilunifiedevaluationinformation}.
Inspired by these approaches, we propose an evaluation methodology for measuring the mutual information between captions and human gaze signals to assess the cognitive plausibility of texts.

\section{Problem Formulation}
We study the problem of aligning video caption generation with human visual attention during video viewing. Our goal is to control caption generation such that the produced descriptions reflect what a human observer attends to while watching a video.
We use gaze signals as a proxy for human cognitive attention, since direct access to humans' internal perceptual or cognitive reasoning is not available. Let $V$ denote a video and $G$ denote the corresponding gaze signal over spatial–temporal regions of $V$.

We formulate gaze-conditioned caption selection as an information-theoretic objective between gaze signals and generated captions. Specifically, we want to find a set of captions (a set of sequences of tokens $\mathbf{t}$) that maximizes the mutual information, denoted by $\mathbf{I}(\cdot)$, between the gaze signal $G$.
\begin{equation}
    \label{eq:mi}
    \max_{\mathbf{t}} \;\; \mathbf{I}(\mathbf{t}; G)
    \;=\;
    \mathbb{E}_{\mathbf{t}, G} \left[ \log \frac{P(\mathbf{t}, G)}{P(\mathbf{t})P(G)} \right],
\end{equation}
where the objective favors captions whose semantics exhibit strong statistical dependence with the viewer's gaze patterns. Here, $P(\mathbf{t}, G)$ denotes the joint distribution over captions and gaze signals, while $P(G)$ and $P(\mathbf{t})$ denote their corresponding marginal distributions under the natural data-generating process. The expectation $\mathbb{E}_{\mathbf{t}, G}[\cdot]$ averages over caption–gaze pairs drawn from the joint distribution $P(\mathbf{t}, G)$, i.e., over the co-occurrences of gaze observations and captions.

\paragraph{Intractability.}
The optimization in (\ref{eq:mi}) is intractable because it requires access to the true marginal distributions $P(G)$ and $P(\mathbf{t})$, which correspond to the natural data-generating processes of human gaze and language. These distributions lack closed-form expressions and cannot be directly evaluated, as they implicitly encode unknown human perceptual and linguistic processes that would require learning full distributions over gaze behavior and caption space, which is essentially the same goal as VLMs and saliency models.

In practice, when we only obtain samples from these distributions, machine learning models approximate the joint behavior through surrogate objectives defined on conditional likelihoods. For example, a pretrained VLM already optimizes a negative log-likelihood objective of the form $-\log P_{\theta}(\mathbf{t} \mid V)$,
which replaces direct modeling of $P(\mathbf{t})$ by learning the conditional distribution of captions given visual input. 
Similarly, alternative approximations for missing marginals often rely on contrastive or variational objectives, such as InfoNCE-style~\cite{oord2018representation} losses in representation learning or KL-regularized objectives in variational inference, which replace intractable mutual information terms with tractable upper or lower bounds.

In the next section, we explain how VEGAS approximates the inaccessible \cref{eq:mi} by leveraging pretrained VLM likelihoods together with gaze-conditioned visual decomposition, thereby avoiding explicit estimation of the intractable marginal distributions.

\section{Video caption Evaluation via GAze Score (VEGAS)}
\label{sec:method}

Our framework operates in two stages.
First, we introduce the \emph{Video caption Evaluation via GAze Score (VEGAS)}, a cross-modal, information-theoretic metric that quantifies how well a caption reflects a viewer's visual attention as indicated by their gaze (Section~\ref{sec:vegas_score}).
Second, we use this score to perform gaze-conditioned rejection sampling over candidate captions, selecting those most aligned with the viewer's attentional focus without any model retraining (Section~\ref{sec:rejection_sampling}).

\subsection{VEGAS: A Gaze-Aware Caption Evaluation Metric}
\label{sec:vegas_score}

\paragraph{Setup and Notation.}
Let $V$ denote the video frames, and let $G \subseteq V$ denote the \emph{gaze-attended} spatial--temporal regions.
Operationally, we construct $G$ by preserving regions near the viewer's fixation in each frame and suppressing regions away from fixation.
The complement $\bar{G}=V\setminus G$ contains the \emph{non-attended} regions.
We assume access to a frozen pretrained VLM that outputs the conditional probability $P_{\theta}(\mathbf{t} \mid V)$ of generating a token sequence given a video input.
Given a candidate caption $c=\mathbf{t}=(t_1,\ldots,t_L)$ of length $L$, we evaluate its token likelihoods under both the original video $V$ and the gaze-attended video $G$.
VEGAS requires no additional training or fine-tuning.

\paragraph{Intuition.}
Intuitively, we approximate the intractable distributions $P(\mathbf{t}, G)$, $P(\mathbf{t})$, and $P(G)$ using the VLM next-token distribution $P_{\theta}(\mathbf{t} \mid V)$, parameterized by the VLM parameters $\theta$. 
A gaze-aligned caption should remain highly predictable from the attended regions $G$ alone, while the non-attended regions $\bar{G}$ should contribute minimal additional information. VEGAS captures this principle by measuring the contribution of non-attended regions to predicting the caption conditioned on the attended regions.

Also, since we cannot compute the expectation in \eqref{eq:mi} directly, we instead use the pointwise mutual information $\mathcal{I}(X, Y)$, defined over per data point as the log-ratio $\log \frac{P(X, Y)}{P(X)P(Y)}$. Unlike mutual information, which averages the pointwise mutual information and is non-negative, pointwise mutual information is unbounded and can take both positive and negative values, reflecting local agreement or disagreement between gaze and caption.

\paragraph{Definition.}
We define \emph{VEGAS} of caption $c$ with respect to gaze $G$ as the pointwise conditional mutual information $\mathcal{I}$ between the caption tokens $\mathbf{t}$ and the non-attended visual regions $\bar{G}$, conditioned on the attended regions $G$:
\begin{equation}
\label{eq:vegas}
    \operatorname{VEGAS}(c, G) 
    \;\triangleq\; 
    \mathcal{I}\bigl(\mathbf{t};\, \bar{G} \mid G\bigr) 
    \;=\; 
    \log \frac{P_{\theta}(\mathbf{t} \mid G,\, \bar{G})}{P_{\theta}(\mathbf{t} \mid G)} 
    \;=\; 
    \log \frac{P_{\theta}(\mathbf{t} \mid V)}{P_{\theta}(\mathbf{t} \mid G)} \quad (\text{Lower is better}),
\end{equation}
where the last equality follows from $G \cup \bar{G} = V$.
Intuitively, a \emph{low} VEGAS score indicates that the caption is well-explained by the gazed regions alone (\textit{i.e.}, it is gaze-aligned), as the token sequence $\mathbf{t}$ is well-predicted by the conditioned gaze $G$ relatively to conditioning on the raw video $V$, and vice versa.
In contrast to \cref{eq:mi}, where higher values indicate stronger gaze alignment, VEGAS captures the information gain in caption prediction when incorporating non-attended regions ($\bar{G} = V \setminus G$), thereby quantifying the extent to which a caption depends on visual content outside the viewer’s focus.
A lower VEGAS score is therefore preferred, as it indicates that non-attended regions contribute little additional predictive information beyond the attended gaze regions, meaning the caption is primarily driven by what the viewer actually looks at.

\paragraph{Token-level Decomposition.}
Because the VLM factorizes the caption likelihood autoregressively, VEGAS decomposes over tokens:
\begin{equation}
\label{eq:vegas_tokens}
    \operatorname{VEGAS}(c, G) 
    \;=\; 
    \sum_{\ell=1}^{L} \log \frac{P_{\theta}(t_\ell \mid t_{<\ell},\, V)}{P_{\theta}(t_\ell \mid t_{<\ell},\, G)}\,,
\end{equation}
which allows us to identify which tokens (and, by extension, which semantic concepts) are and are not supported by the viewer's gaze.
Namely, it captures the lift in the VLM’s per-token confidence when inputting on the full video $V$ versus inputting on the video with the non-attended regions $\bar{G}$ removed.

\paragraph{Properties.}
VEGAS is (i)~\emph{cross-modal}, requiring no human-written reference caption as gold standard and jointly grounding language and vision through the VLM's internal distribution; and (ii)~\emph{personalized}, as it evaluates a caption with respect to a specific individual's gaze pattern.

\subsection{Gaze-Conditioned Rejection Sampling}
\label{sec:rejection_sampling}

Given a pretrained VLM and a viewer's gaze signal $G$, our goal is to produce a caption that is gaze-aligned without retraining the model.
We achieve this via rejection sampling over the VLM's own caption distribution.

We first draw $N$ candidate captions $\{c_1, \ldots, c_N\}$ by sampling from one or multiple VLMs with various temperature and reasoning settings conditioned on the full video $V$.
For each candidate $c_i$, we compute $\operatorname{VEGAS}(c_i, G)$ using~\cref{eq:vegas_tokens}.
We then select the caption with the \emph{lowest} VEGAS score:
\begin{equation}
\label{eq:rejection}
    c^{*} \;=\; \arg\min_{c_1,\ldots, c_N} \;\operatorname{VEGAS}(c_i, G), \; \forall i \in [1, N] \quad (\text{VEGAS-based Rejection Sampling}).
\end{equation}
The problem in (\ref{eq:rejection}) selects the caption whose content is best explained by the viewer's attended regions, effectively filtering out captions that describe unattended content.

\paragraph{Practical Advantages.}
Selecting captions with VEGAS has several practical advantages.
First, it requires \emph{no retraining or fine-tuning} of the VLM; gaze is used at inference time.
Second, because the base distribution is the VLM's own caption prior, selected captions inherit the fluency and coherence of the pretrained model.
Third, the framework is agnostic to the specific VLM architecture, making it applicable to any model that provides token-level likelihoods conditioned on visual input, even online APIs like GPT and Gemini.

\section{Dataset Curation}
We curate a multimodal dataset for gaze-conditioned caption selection and evaluation, pairing visual content, synchronized human gaze, and free-form captions to capture distinctive attention-language alignments.
Our curation spans two distinct domains: (1) egocentric daily activities using the Aria Everyday Activities (AEA) dataset, and (2) instructional slide presentations using SlideVQA. Participant instructions and user study interfaces are detailed in Appendix~\ref{app:instructions} and \ref{app:user_study_interfaces}.

We recruited human annotators via Prolific~\citep{prolific2026}, restricting eligibility to individuals aged 18 years or older who reside in the United States and are fluent in English. To ensure accurate data collection, participation was restricted to desktop computer users. For the SlideVQA user study, participants were additionally required to use a webcam to enable remote eye tracking.

\subsection{Egocentric Video Captioning Dataset (Aria Everyday Activities)}
\label{sec:aea_dataset}

The AEA subset captures gaze-conditioned language in dynamic, real-world environments. Project Aria smart glasses provide the first-person video streams alongside hardware-synchronized human gaze measurements natively recorded from the wearer.

\paragraph{Video Segmentation.}
We first segment raw egocentric recordings into shorter, action-centric clips to isolate discrete tasks. We leverage \texttt{Gemini-3.1-Pro-Preview} to identify boundaries, prompting the model to output precise start and end timestamps along with a short, first-person summary beginning with ``I...'' in the simple present tense. This minimizes semantic ambiguity and establishes a tighter temporal mapping between visual attention and language.
The Gemini-generated segmentation summaries are used only for boundary proposals. They are not shown to human annotators and has no effect on human caption quality.

\paragraph{Video Clip Selection.}
Our goal is to evaluate on gaze-sensitive clips where personalization is expected to matter, namely segments where access to gaze substantially changes the generated caption.
To identify such segments, we run captioning inference with \texttt{Gemini-3.1-Pro-Preview} under two different settings: (1) using gaze-conditioned inputs, where we mask frame regions centered at fixation points, and (2) using the raw video without gaze information.
We measure the semantic divergence between these pairs using SBERT similarity and evaluate them via the VEGAS metric using \texttt{Gemma-4-31B-IT}. We retain the top $25\%$ of segments exhibiting the lowest SBERT similarity and highest VEGAS scores, yielding $332$ highly gaze-dependent video clips for human annotation.

\paragraph{Caption Annotation.}
Using the Qualtrics XM Platform~\citep{qualtrics_xm_2024}, recruited annotators viewed each segmented clip overlaid with a synchronized gaze visualization (indicated by a red dot). We instructed participants to write a single-sentence caption exceeding $30$ characters that explicitly describes the action emphasized by the wearer's visual attention. This protocol enforces target-driven descriptions rather than a passive, exhaustive narration of the entire background scene.

\subsection{Slide Deck Captioning Dataset (SlideVQA)}
The SlideVQA subset contains instructional and educational slide decks spanning diverse topics and visual layouts. 
Since SlideVQA does not provide gaze annotations, we collect viewer attention data through a controlled viewing study.
We select $30$ slide decks from the dataset and present the first $10$ slides of each deck to participants for captioning and gaze recording. We recruit $5$ participants per slide deck. Each participant views a slide sequence and provides a caption per slide while we record their gaze behavior.

\paragraph{Data Collection.}
We recorded participant gaze trajectories using the webcam-based eye-tracking model on the RealEye platform~\citep{realeye2025}. The platform natively processes visual streams locally on the user's device, capturing and storing only anonymized, timestamped coordinate vectors without saving any raw camera video. Participants were instructed to view each slide naturally for a minimum of $20$ seconds before proceeding to the annotation phase.
Directly after viewing a slide, participants compose a free-form caption between $30$ and $200$ characters detailing the prominent text blocks, visual elements, or core semantic concepts that anchored their attention. 

\paragraph{Data Processing.}

Gaze coordinates were normalized relative to the native slide resolutions and aggregated into continuous spatial attention maps. RealEye provides operational quality scores ranging from $1$ (\textit{Very Low}) to $6$ (\textit{Perfect})\footnote{RealEye assigns a quality grade from 1 (\textit{Very Low}) to 6 (\textit{Perfect}) based on eye-tracking sampling rate, fixation computability, percentage of items with gaze data, total gaze data length, and gaze-on-screen coverage. Grades $\geq 3$ (\textit{Average} and above) achieve a sampling rate of at least $10$~Hz and $\geq 30\%$ gaze-on-screen coverage.}. 
For our experiments, we enforce two quality filters: (1) we exclude participants with a RealEye quality grade below $3$, and (2) we purge participants who submitted duplicate text responses across different slides within a single deck. Ultimately, $142$ out of $150$ total participants were retained, yielding an average calibration score of $4.88$.

\subsection{Dataset Statistics}

Our final curated dataset comprises $632$ total visual samples paired with $2{,}981$ high-quality, human-written captions (see Table~\ref{tab:dataset_stats}). The egocentric AEA subset ($332$ action clips derived from $91$ unique parent videos, totaling $46.78$ minutes) captures everyday tasks like cooking, computing, and object manipulation. The presentation-based SlideVQA subset ($300$ individual slides across $30$ decks) spans dense academic, business, and scientific layouts. 
Captions average $14.8 \pm 5.6$ words for AEA and $16.0 \pm 7.1$ words for SlideVQA. The notable variation in reading behaviors and linguistic styles between the two subsets underscores the value of cross-domain gaze-aware evaluation.

\begin{table}[h]
\centering
\caption{Statistics for the Aria Everyday Activities (AEA) and SlideVQA sub-datasets. Caption statistics are reported for human-written annotations (after quality filtering for SlideVQA).}
\label{tab:dataset_stats}
\begin{tabular}{lcc}
\toprule
\textbf{Metric} & \textbf{AEA (Egocentric)} & \textbf{SlideVQA (Presentation)} \\ \midrule
Number of Parent Source Units (Videos / Decks) & 91 & 30 \\
Number of Annotated Samples (Clips / Slides) & 332 & 300 \\
Total Retained Human Captions & 1,660 & 1,321 \\
Mean Caption Length (Words) & $14.8 \pm 5.6$ & $16.0 \pm 7.1$ \\
\bottomrule
\end{tabular}
\end{table}

\subsection{Ethical Considerations}
We recognize that gaze trajectories can reveal sensitive information regarding individual cognitive patterns and implicit user intent. 
For SlideVQA, participants were informed during consent that gaze would be collected and reused for research. 
We will publish both raw gaze traces and processed attention heatmaps to enable reproducibility.
To mitigate privacy risks, all personally identifiable information (PII) in metadata and annotations was stripped during post-processing.
For AEA, the gaze data are used directly from the source dataset, whose videos were already processed with EgoBlur~\citep{raina2023egoblur} to automatically detect and obscure PII such as faces and license plates.
All data collection and annotation procedures are approved by the Institutional Review Board (IRB) and follow institutional ethical guidelines.

\section{Experiments}
\label{sec:experiments}

\paragraph{Candidate Caption Generation.}
We generate candidate video captions using multimodal large language models spanning closed-source models (\texttt{Gemini-3.1-Pro-Preview} and \texttt{GPT-5.5}) and an open-source model (\texttt{Qwen3.6-35B-A3B}).
All models use the same candidate-captioning prompt, provided in Appendix~\ref{app:prompts}. Candidate generation uses original visual inputs: for AEA, we sample frames from the original extracted clip videos at $1$ fps; for SlideVQA, we provide the slide images directly. For each sample, we generate one caption from each model at decoding temperatures in $\{0.0, 1.0\}$, yielding up to $N=6$ candidates per sample before post-processing (see Appendix~\ref{app:caption_postprocessing}).
We choose these models because they are among the strongest performers on multimodal understanding tasks and provide a diverse pool of caption candidates.

\paragraph{VEGAS Scoring.}
For VEGAS evaluation, we deploy \texttt{Gemma-4-31B} locally to score captions in a controlled setting, avoiding API-side model updates and serving variability.
We set the model temperature to $T=1$ to score captions under the model's native, untempered next-token distribution.
The random seed only affects token sampling, not the underlying next-token probabilities, and therefore does not alter the model's intrinsic likelihood estimates.
Per \cref{eq:vegas_tokens}, throughout the experiments, we retain only the top-20 token probabilities at each decoding step and use a fallback log-probability of $-30$ for tokens outside this set.
Appendix~\ref{app:vegas_implementation_robustness} ablates these choices and an alternative Qwen2.5-VL-32B-Instruct scorer, showing that downstream SBERT and retrieval performance remain stable across the implementation grid.
Additional details on compute resources and caption post-processing are provided in Appendix~\ref{app:compute_resources} and \ref{app:caption_postprocessing}, respectively.

\paragraph{Semantic Alignment Protocol.}
We evaluate whether lower-VEGAS captions are semantically closer to human annotations using SBERT similarity to each sample's human-caption centroid.
First, we compare naive Gemini captions against VEGAS-selected captions and test paired similarity differences with a Wilcoxon signed-rank test. For AEA, the naive Gemini caption is the first-person clip summary produced during video segmentation; for SlideVQA, it is the Gemini caption generated directly from the slide image.
We then control for candidate-pool quality by comparing selectors within the same caption pool: random selection, raw VEGAS, length-normalized VEGAS, and a best-in-pool oracle. 
The length-normalized score is defined as $\operatorname{VEGAS}_{\mathrm{norm}}=\operatorname{VEGAS}/L$, where $L$ is the number of scored tokens, and the best-in-pool oracle selects the caption with highest SBERT similarity to the human-caption centroid.
Finally, we run two diagnostics: a gaze-corruption ablation to test whether VEGAS depends on the correct gaze signal, and a caption-length analysis to test whether improvements are explained by shorter captions.

\paragraph{Caption-to-Video Retrieval Protocol.}
To evaluate whether VEGAS turns wearer attention into a better semantic index, we perform caption-to-video retrieval on AEA.
Each query is a human-written caption describing what the camera wearer attended to in a clip, and the retrieval index is constructed from captions automatically generated from the candidate videos. 
The candidate captions are pooled across three VLMs (\texttt{Gemini-3.1-Pro-Preview}, \texttt{GPT-5.5}, and \texttt{Qwen3.6-35B-A3B}) and two decoding temperatures ($T=0$ and $T=1$). 
Candidate captions are pooled across the three VLMs and two decoding temperatures.
We compare random VLM caption indexing, VEGAS-selected VLM caption indexing, and a human pairwise reference in which one human caption queries another human caption for the same video.
Retrieval metrics are averaged over human-query runs and reported as Precision@$K$, Recall@$K$, and mAP@$K$.

\section{Results}
\label{sec:results}

\paragraph{How diverse are human captions for identical visual content?}
To quantify human annotation diversity, we embed the human captions for each AEA clip using \texttt{sentence-transformer/all-mpnet-base-v2}~\cite{song2020mpnet}. For each clip, we fit a local 3-dimensional PCA to the human-caption embeddings, center the coordinates at the human-caption centroid, and pool the standardized coordinates across clips. We use the same encoder for all SBERT-based analyses in this section.
\begin{wrapfigure}{r}{0.4\textwidth}
    \vspace{-1em}
    \centering
    \includegraphics[width=0.4\textwidth, trim=0 50 0 20, clip]{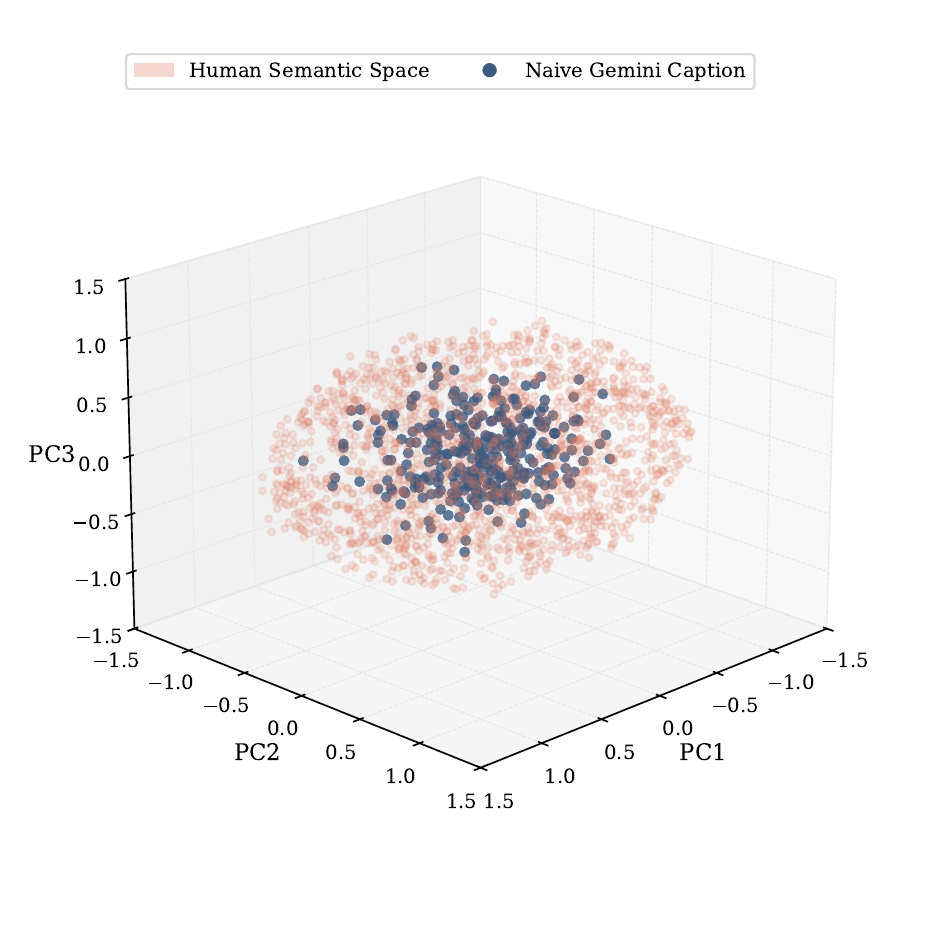}
    \vspace{-2em}
    \caption{\textbf{PCA visualization.}
    Human captions are much more diverse than naive Gemini captions in semantic space.
    }
    \label{fig:human_diversity}
    \vspace{-1em}
\end{wrapfigure}
Across $332$ AEA clips, the local PCA retains $89.29\%$ of human-caption variance on average. In this PCA space, the mean pairwise human-human distance is $1.53$, compared with a mean Gemini-human distance of $1.07$ and a mean Gemini--centroid distance of $0.48$.
As shown in \Cref{fig:human_diversity}, naive Gemini captions lie close to the human consensus centroid but cover only a small portion of the semantic variation among human annotations. This indicates that human captions for the same visual content are substantially more diverse than generic VLM captions, motivating personalized captioning methods that can reflect individual perceptual focus.

\begin{figure}[h]
    \centering
    \includegraphics[width=\linewidth, trim=0 0 0 40, clip]{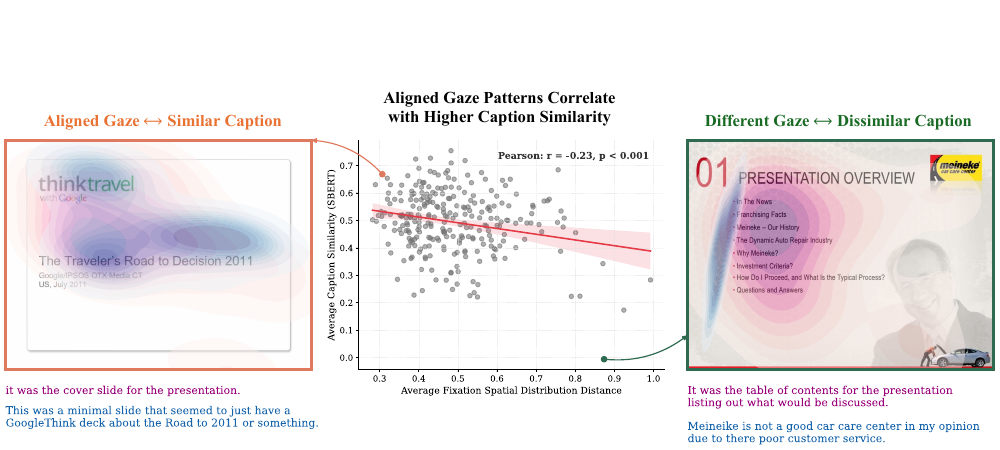}
    \caption{\textbf{Gaze-caption personalization in SlideVQA.} The center panel displays a scatter plot of average caption semantic similarity (SBERT) against average fixation spatial distribution distance between participant pairs. The left and right panels present qualitative slides where two participants (one in pink, the other in blue) exhibit distinct visual gaze distributions on identical slides.
    }
    \label{fig:slidevqa_personalization}
\end{figure}

\paragraph{Are individual differences in gaze patterns reflected in human captions?}
Given this high diversity, we evaluate whether individual spatial gaze patterns can act as a reasonable proxy to guide personalized captioning. In \Cref{fig:slidevqa_personalization} (center panel), the x-axis represents the average fixation spatial distribution distance between participant pairs, and the y-axis represents the average SBERT caption similarity. 
We observe a significant negative Pearson correlation ($r = -0.23, p < 0.001$), indicating that as visual gaze patterns diverge spatially, their written captions also diverge semantically. 
This directly demonstrates that individual visual attention patterns are significantly but partially associated with their semantic interpretation of visual content, validating gaze as a reasonable proxy for personalized human captions.

\begin{figure}[h]
    \centering
    \includegraphics[width=0.8\linewidth, trim=20 0 20 0, clip]{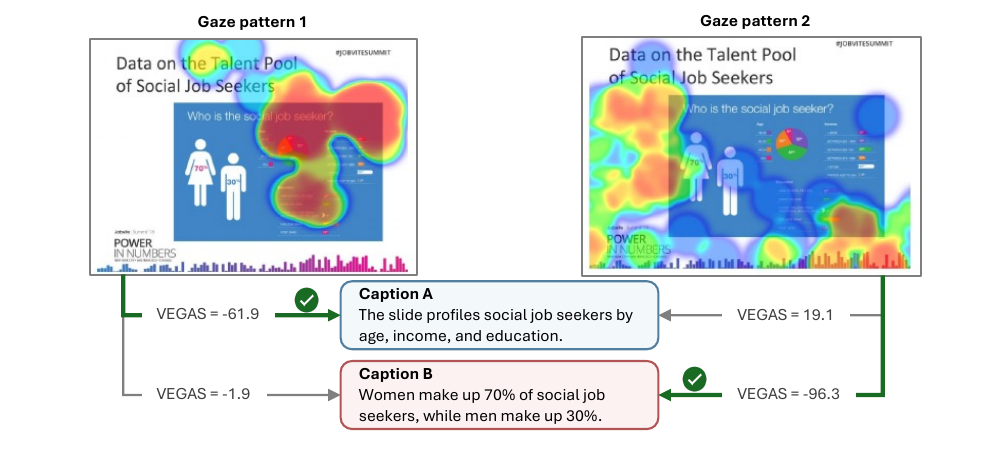}
    \caption{\textbf{VEGAS scoring in SlideVQA.} For the same slide and candidate captions, VEGAS assigns lower scores to captions whose tokens align with each viewer's attended regions. 
    }
    \label{fig:slidevqa_personalized_social_job_seekers}
    \vspace{-1em}
\end{figure}

Our method uses this gaze-language association to personalize caption selection. \Cref{fig:slidevqa_personalized_social_job_seekers} shows a qualitative example where the visual input is fixed, but the gaze pattern changes; VEGAS correspondingly assigns lower scores to different candidate captions whose tokens better match the regions attended by each viewer.

\begin{figure}[h]
    \centering
    \includegraphics[width=0.8\linewidth]{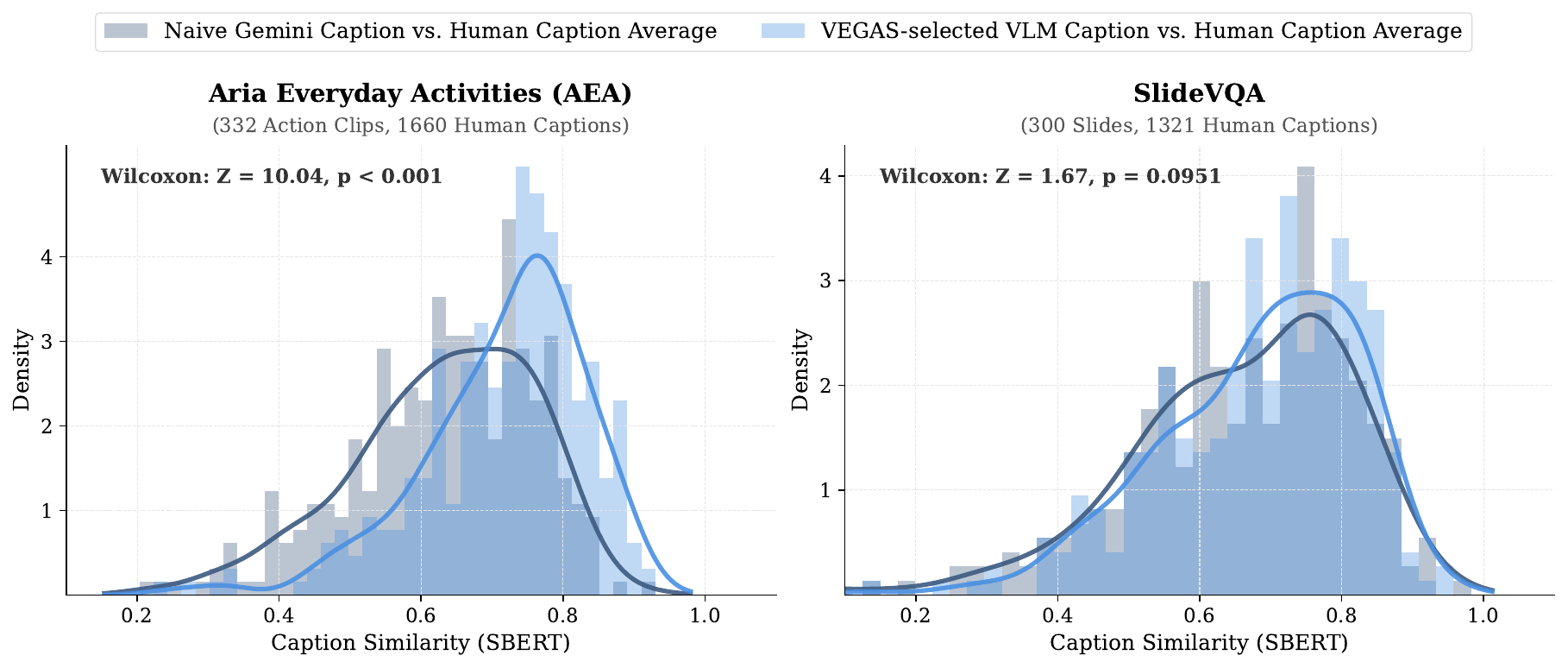}
    \caption{\textbf{Density distribution of SBERT similarity.}
    The x-axis denotes caption semantic similarity (SBERT) relative to the average human caption, and the y-axis shows empirical probability density. We compare naive Gemini captions (dark blue) against VEGAS-selected captions (light blue) from the pooled \texttt{Gemini-3.1-Pro-Preview}, \texttt{GPT-5.5}, and \texttt{Qwen3.6-35B-A3B} candidates. VEGAS selection shifts mean SBERT similarity by +0.0856 (+13.53\%) on AEA and +0.0256 (+3.88\%) on SlideVQA.
    }
    \label{fig:sbert}
\end{figure}

\paragraph{Are captions with lower VEGAS semantically closer to human captions?}
For AEA, the naive Gemini captions are the first-person clip summaries produced during the video-segmentation step in \Cref{sec:aea_dataset}. 
As shown in \Cref{fig:sbert}, captions selected by minimizing VEGAS (light blue) shift toward higher SBERT similarity relative to these naive Gemini captions (dark blue), increasing mean similarity by $+0.0856$ ($+13.53\%$) on the $[-1, 1]$ SBERT scale.
This rightward shift is statistically significant under a Wilcoxon signed-rank test ($Z = 10.04, p < 0.001$), indicating that VEGAS selection improves semantic alignment over generic segmentation summaries in egocentric activity videos, where gaze helps identify the actions and objects most relevant to human annotators.
On SlideVQA, VEGAS selection produces a smaller positive mean shift of $+0.0256$ ($+3.88\%$), but this difference is not statistically significant ($Z = 1.67, p = 0.0952$).
We therefore interpret the SlideVQA result cautiously: because much of the slide text is directly visible, generic VLM captions can already match one plausible human description, while different annotators may still emphasize different valid aspects of the same slide. This makes SlideVQA a noisier target for single-reference semantic similarity and reinforces the value of human annotations that preserve diverse, attention-driven interpretations.

We next compare selectors within the same VLM candidate pool (\Cref{tab:sbert_candidate_pool_selection_summary}). On AEA, raw VEGAS improves over random selection by $+0.013$ SBERT ($p<0.05$), while length-normalized VEGAS shows a smaller positive trend. On SlideVQA, VEGAS is tied with random selection, consistent with slides admitting multiple valid semantic interpretations and providing a weaker gaze-conditioned signal. The best-in-pool oracle remains substantially higher in both domains, indicating room for stronger future selection objectives.

\begin{table}[h]
    \centering
    \caption{\textbf{Selection from the same candidate pool.} Each selector chooses one caption per item from the same pool. SBERT is mean $\pm$ standard deviation to the human-caption centroid. $\Delta$ is relative to Random; significance uses a paired Wilcoxon test ($^{*}p<0.05$, $^{**}p<0.01$).}
    \label{tab:sbert_candidate_pool_selection_summary}
    \resizebox{0.6\linewidth}{!}{%
    \begin{tabular}{lllll}
        \toprule
        Domain & $n$ & Selector & SBERT & $\Delta$ vs Random \\
        \midrule
        \multirow{4}{*}{AEA} & \multirow{4}{*}{$332$}
        & Random (seed=42) & $0.705 \pm 0.124$ & -- \\
        & & VEGAS & $0.718 \pm 0.115$ & $+0.013^{*}$ \\
        & & VEGAS / token & $0.714 \pm 0.117$ & $+0.009$ \\
        & & Best-in-pool & $0.800 \pm 0.075$ & $+0.095^{**}$ \\
        \midrule
        \multirow{4}{*}{SlideVQA} & \multirow{4}{*}{$300$}
        & Random (seed=42) & $0.685 \pm 0.139$ & -- \\
        & & VEGAS & $0.685 \pm 0.138$ & $-0.000$\\
        & & VEGAS / token & $0.685 \pm 0.137$ & $+0.000$ \\
        & & Best-in-pool & $0.742 \pm 0.114$ & $+0.057^{**}$ \\
        \bottomrule
    \end{tabular}%
    }
\end{table}

\begin{figure}[h]
    \centering
    \includegraphics[width=0.9\linewidth, trim=0 80 0 0, clip]{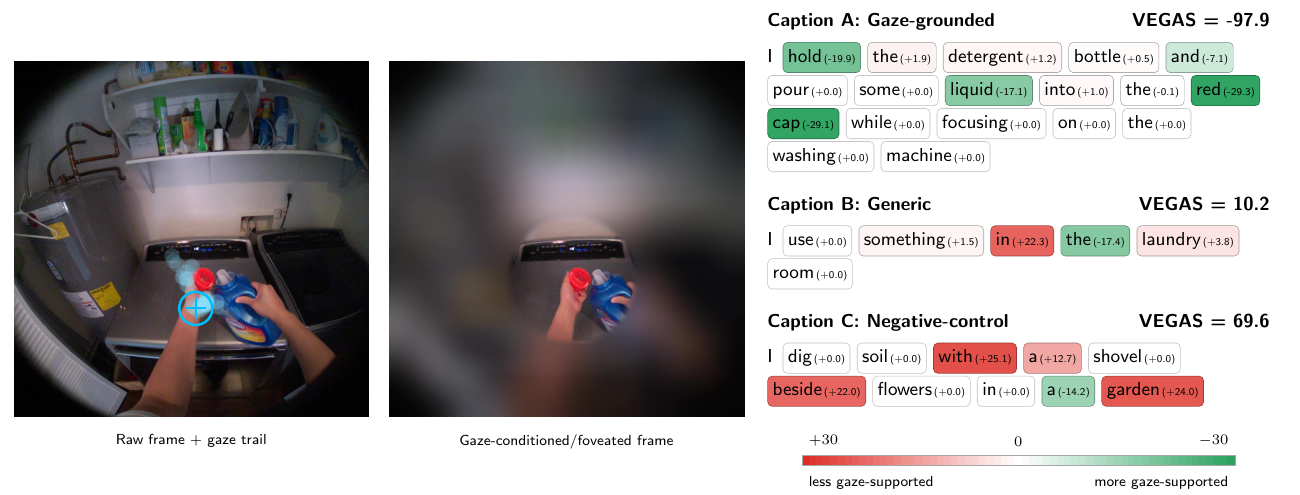}
    \caption{\textbf{AEA qualitative example.} VEGAS assigns the lowest score to the caption aligned with the attended object, compared with a generic caption and an unrelated negative control. Token scores show that terms tied to gaze-attended regions such as ``liquid'' and ``red cap'' in Caption A tend to be more supported (green), while unrelated terms like ``garden'' in Caption C are less supported (red).}
    \label{fig:aea_detergent_cap}
\end{figure}

To illustrate VEGAS at the token level, \Cref{fig:aea_detergent_cap} shows an AEA example where the attended object disambiguates the best caption from generic and unrelated alternatives. An additional example is provided in \Cref{fig:aea_tv_screen_appendix} of Appendix~\ref{app:qualitative_examples}.

In the same appendix section, we also include two SlideVQA bad cases in \Cref{fig:slidevqa_bad_cases}, where reasonable captions receive high VEGAS scores. 
These examples help explain the weaker quantitative gains on SlideVQA: unlike AEA, where captions often depend on grounding concrete objects, slide captioning frequently requires conceptual abstraction. Gaze indicates where a viewer looked, but may not fully determine how the viewer interprets chart trends, causal relationships, or high-level concepts.

\paragraph{Is VEGAS simply selecting shorter captions?}
Because VEGAS is computed as a token-level sum, a natural concern is that it may simply prefer shorter or more generic captions rather than captions that better align with gaze.
We therefore run a post-hoc diagnostic over the full VLM candidate pools used for rejection sampling.
We correlate VEGAS with caption length, measured by word count, and caption specificity, approximated by POS-tag counts such as nouns, verbs, adjectives, numeric/entity-like tokens, and content words.
As shown in \Cref{tab:length_specificity_correlations}, both correlations are near zero in AEA and SlideVQA, providing no evidence that VEGAS is driven by brevity or genericness. 
This can also be seen in \Cref{tab:sbert_candidate_pool_selection_summary} where raw VEGAS improves SBERT over random selection ($+0.013$, $p=0.040$), while length-normalized VEGAS shows a smaller, non-significant gain ($+0.009$, $p=0.183$) in AEA, suggests that improvements are not explained by caption length alone.

\begin{table}[h]
    \centering
    \caption{\textbf{Caption length and specificity vs. VEGAS.} Pearson correlations over raw VLM candidates show near-zero association with VEGAS in both domains.}
    \label{tab:length_specificity_correlations}
    \resizebox{0.5\textwidth}{!}{%
    \begin{tabular}{lrrrr}
        \toprule
        Domain & $n$ & Word count $r$ ($p$) & Specificity $r$ ($p$) \\
        \midrule
        AEA & $1814$ & $0.001$ ($0.979$) & $0.026$ ($0.275$) \\
        SlideVQA & $1578$ & $-0.014$ ($0.580$) & $-0.014$ ($0.579$) \\
        \bottomrule
    \end{tabular}
    }
\end{table}

\paragraph{Is VEGAS sensitive to gaze corruption?}
To test whether VEGAS is sensitive to the correct gaze signal, we compare each AEA caption's score under true gaze against two corrupted gaze conditions: randomly mismatched gaze from another clip and a center-bias gaze surrogate.
As shown in \Cref{tab:aea_gaze_corruption_ablation}, random gaze increases VEGAS by $+1.786$ points on average relative to true gaze, with a $95\%$ bootstrap confidence interval of $[+0.365,+3.289]$ and a paired Wilcoxon test of $p=0.040$.
This indicates that mismatching a caption with another clip's gaze measurably worsens gaze-caption alignment.
Center-bias gaze also increases the mean score by $+1.412$ points, but the confidence interval crosses zero ($[-0.181,+3.019]$) and the Wilcoxon test is marginal ($p=0.058$).
Overall, true AEA gaze contains measurable information beyond a randomly mismatched gaze baseline, while the weaker separation from center-bias gaze suggests partial overlap between true gaze and coarse central salience in egocentric video.

\begin{table}[h]
    \centering
    \caption{\textbf{Gaze corruption ablation on AEA.} Lower VEGAS is better. $\Delta$ denotes the paired score difference relative to true gaze, $\operatorname{VEGAS}(c,G_{\mathrm{condition}}) - \operatorname{VEGAS}(c,G_{\mathrm{true}})$. Positive $\Delta$ therefore indicates worse gaze-caption alignment than the true gaze condition.}
    \label{tab:aea_gaze_corruption_ablation}
    \resizebox{0.7\textwidth}{!}{%
    \begin{tabular}{lrrrrr}
        \toprule
        Condition & $n$ & Mean VEGAS & Mean $\Delta$ & $95\%$ CI for $\Delta$ & Wilcoxon $p$ \\
        \midrule
        True gaze & $327$ & $47.632$ & -- & -- & -- \\
        Random gaze & $327$ & $49.419$ & $+1.786$ & $[+0.365,+3.289]$ & $0.040$ \\
        Center-bias gaze & $327$ & $49.044$ & $+1.412$ & $[-0.181,+3.019]$ & $0.058$ \\
        \bottomrule
    \end{tabular}
    }
\end{table}

\paragraph{Does VEGAS improve downstream caption-to-video retrieval?}
To evaluate whether VEGAS turns wearer attention into a better semantic index, we perform caption-to-video retrieval on AEA.
Each query is a human-written caption describing what the camera wearer attended to in a clip, and the retrieval index is constructed from captions automatically generated from the candidate videos.
The candidate captions are pooled across three VLMs (\texttt{Gemini-3.1-Pro-Preview}, \texttt{GPT-5.5}, and \texttt{Qwen3.6-35B-A3B}) and two decoding temperatures ($T=0$ and $T=1$).

We compare three index-construction strategies.
The first uses a randomly selected VLM caption from the pooled captions.
The second, \textsc{VEGAS (Ours)}, selects the lowest-VEGAS caption from the same pool.
The third is a human pairwise reference: one human caption is used as the query, and a different human caption for the same video is used as the single human-authored index entry.
Results are averaged over all ordered pairs of distinct annotators, measuring human-human retrieval agreement.

As shown in \Cref{tab:human_query_retrieval_results}, \textsc{VEGAS (Ours)} consistently improves over random VLM captions across all reported precision, recall, and mAP metrics.
The gains are modest at rank 1 ($+1.14\%$ for Precision/Recall/mAP@1) but become more pronounced at larger retrieval depths, improving Recall@5 by $+4.16\%$, Recall@10 by $+4.10\%$, mAP@5 by $+2.48\%$, and mAP@10 by $+2.46\%$.
Measured as random-to-human gap closure, VEGAS recovers $15.0\%$ of the Rank@1 gap, but a substantially larger share at broader retrieval depths: $40.8\%$ for Recall@5, $42.3\%$ for Recall@10, $27.8\%$ for mAP@5, and $27.8\%$ for mAP@10.
This pattern suggests that VEGAS most strongly improves the semantic neighborhood of the retrieved results: selected captions often move the correct video into the top few candidates, even when they are not yet discriminative enough to make it the top-ranked result.
These results suggest that lower-VEGAS captions are better aligned with the descriptions humans naturally use to search for videos.
The human pairwise row shows the remaining gap between machine-selected and human-written index captions.

\begin{table}[h]
    \centering
    \caption{\textbf{Human-query retrieval results on AEA.} We encode human annotations with SBERT and average retrieval metrics over human-query runs. Candidate indices contain one caption per video: a random VLM caption, the lowest-VEGAS caption selected from pooled VLM captions (\texttt{Gemini-3.1-Pro-Preview}, \texttt{GPT-5.5}, and \texttt{Qwen3.6-35B-A3B}), or a human caption in an ordered pairwise setup where annotator $i$ queries annotator $j$ with $i \neq j$. \textsc{VEGAS (Ours)} improves over random VLM captions across metrics.}
    \label{tab:human_query_retrieval_results}
    \vspace{1em}
    \resizebox{\textwidth}{!}{%
    \begin{tabular}{l ccc ccc ccc}
    \toprule
    \textbf{Candidate Source} &
    \multicolumn{3}{c}{\textbf{Precision@K}} &
    \multicolumn{3}{c}{\textbf{Recall@K}} &
    \multicolumn{3}{c}{\textbf{mAP@K}} \\
    \cmidrule(lr){2-4} \cmidrule(lr){5-7} \cmidrule(lr){8-10}
    & @1 & @5 & @10 & @1 & @5 & @10 & @1 & @5 & @10 \\
    \midrule
    Random VLM
    & 22.41\% & 9.67\% & 6.03\%
    & 22.41\% & 48.37\% & 60.30\%
    & 22.41\% & 31.73\% & 33.31\% \\

    \textsc{VEGAS (Ours)}
    & \textbf{23.55\%}~{\scriptsize (+1.14\%)} & \textbf{10.51\%}~{\scriptsize (+0.84\%)} & \textbf{6.44\%}~{\scriptsize (+0.41\%)}
    & \textbf{23.55\%}~{\scriptsize (+1.14\%)} & \textbf{52.53\%}~{\scriptsize (+4.16\%)} & \textbf{64.40\%}~{\scriptsize (+4.10\%)}
    & \textbf{23.55\%}~{\scriptsize (+1.14\%)} & \textbf{34.21\%}~{\scriptsize (+2.48\%)} & \textbf{35.77\%}~{\scriptsize (+2.46\%)} \\

    \midrule
    Human Pairwise
    & 30.02\% & 11.71\% & 7.00\%
    & 30.02\% & 58.57\% & 69.98\%
    & 30.02\% & 40.64\% & 42.19\% \\
    \bottomrule
    \end{tabular}
    }
\end{table}

\section{Limitations}

VEGAS inherits the limitations of the underlying VLMs, including hallucinations, perceptual errors, and imperfect token likelihood calibration. If a VLM produces similar hallucinated interpretations with and without gaze conditioning, VEGAS may assign overly favorable scores.
A second limitation is that gaze reveals where a viewer attends, but not always how they interpret what they see. This matters for domains where captioning may require conceptual abstraction, information aggregation, or reasoning across visual elements. Thus, VEGAS may be most effective when attention is closely tied to concrete objects, actions, or entities.
Finally, VEGAS requires gaze annotations. Although webcam-based eye tracking reduces collection cost, it still requires controlled data acquisition and participant involvement. In deployed systems, explicit gaze may be available in smart glasses, while implicit attention proxies may be more realistic for web and video platforms.

\section{Conclusion}

Existing VLMs generate video captions that largely ignore individual viewer attention, despite the fact that human video descriptions are inherently diverse and closely tied to where people look. Motivated by this observation, we introduce VEGAS, a cross-modal, information-theoretic metric that measures how well a video caption aligns with a viewer’s gaze, along with a VEGAS-based rejection sampling method that uses gaze as a proxy for viewer intent to select attention-aware captions from candidate VLM outputs. This approach operates entirely at inference time and requires no retraining of the underlying vision-language model. By re-ranking and selecting among candidate captions generated by existing VLMs, VEGAS turns viewer attention into a better semantic index for downstream retrieval and attention-aware interaction.

To evaluate this framework, we curate a multimodal dataset pairing gaze trajectories, videos, and human-written captions across egocentric and instructional slide domains. 
On AEA, VEGAS-selected captions significantly improve semantic alignment with human annotations and yield modest but consistent gains in downstream caption-to-video retrieval.
On SlideVQA, the gains are smaller and not statistically significant, suggesting that gaze is a weaker selector when captions require conceptual abstraction rather than grounding concrete attended objects or actions.
Together, these results show that test-time gaze can provide a practical mechanism for viewer-aware caption selection, while also highlighting that its benefits depend on how directly visual attention maps onto semantic interpretation.

\paragraph{Future work.} 
Looking ahead, VEGAS opens several directions. Future work could improve scalability by replacing explicit gaze with saliency models, interaction traces, or other implicit attention proxies, and by extending VEGAS to longer videos and real-time personalization. Robustness could be improved through external verification, grounding constraints, or calibrated scoring models that reduce sensitivity to VLM hallucinations and likelihood miscalibration. VEGAS could also be distilled into captioning models or reformulated as a differentiable training objective, enabling applications such as personal memory retrieval, assistant question answering, and learning-support systems.

\section*{Acknowledgement}
We thank Kristen Grauman and Yu Chen for their valuable feedback and helpful discussions.

\newpage
\bibliography{ref}
\bibliographystyle{tmlr}

\newpage
\appendix
\phantomsection
\pdfbookmark[0]{Appendix}{app:appendix}
\section*{Appendix}

\section{Human Study Instructions}
\label{app:instructions}
All participants provide informed consent prior to participation and receive compensation following Prolific payment guidelines and institutional policies. The user interface is presented in

\begin{tcolorbox}
    [ 
    enhanced,
    breakable,
    colframe=datasetCcolor!70!black, colback=datasetCcolor!10!white, colbacktitle=datasetCcolor!70!black,
    coltitle=white, title=\textbf{Instructions for \texttt{SlideVQA}}, sharp
    corners=south, boxrule=0.8mm, width=\textwidth, enlarge left by=0mm, enlarge right by=0mm ] 
    \footnotesize
    You will view a deck of 10 slides and summarize the content.
    
    \textbf{Before starting}
    \begin{itemize}
        \item You will complete a brief gaze calibration to estimate where you are looking.
        \item No images or video of your face are stored—only gaze coordinates.
        \item Keep your head relatively still during calibration and the task.
    \end{itemize}
    If you are using an external monitor, open the study on the device with the camera.
    
    \textbf{During the study}
    \begin{itemize}
        \item Read each slide at your normal pace.
        \item After each slide, write a short paragraph summarizing the main ideas and key takeaways.
    \end{itemize}
    
    \textbf{Requirements}
    \begin{itemize}
        \item Age 18+
        \item Fluent in English
    \end{itemize}

    \textbf{Study Instruction:}
    You will view a deck of 10 slides.
    \begin{itemize}
        \item Please read each slide carefully at your normal pace.
        \item After each slide, write a short paragraph summarizing the main ideas/takeaways of the slide.
    \end{itemize}

    \textbf{Important}
    \begin{itemize}
        \item Write meaningful summaries based on the slide content. Submissions may be rejected if responses are nonsensical or repetitive.
    \end{itemize}

    \textbf{Instructions after each slide:}
    
    Write one clear, natural paragraph summarizing the main ideas/takeaways of the last slide.
\end{tcolorbox}

\begin{tcolorbox}[
    enhanced,
    breakable,
    colframe=datasetAcolor!70!black, %
    colback=datasetAcolor!10!white, 
    colbacktitle=datasetAcolor!70!black,
    coltitle=white, 
    title=\textbf{Instructions for \texttt{Aria Everyday Activities}}, 
    sharp corners=south, 
    boxrule=0.8mm, 
    width=\textwidth, 
    enlarge left by=0mm, 
    enlarge right by=0mm
] 
\footnotesize 
You are invited to take part in a research study on first-person activity recognition. You will watch short videos recorded from smart glasses and describe the wearer's actions.

    \textbf{What You'll Do}
    \begin{itemize}
        \item Watch ~27-35 short video clips ($\sim$5 minutes total)
        \item Write one sentence per clip describing the action
    \end{itemize}
    
    \textbf{Requirements}
    \begin{itemize}
        \item Age 18+
        \item Fluent in English
    \end{itemize}

    \textbf{Purpose}
    
    You are invited to participate in a research study on first-person activity recognition. You will watch short video clips recorded from smart glasses and write captions describing the wearer’s actions.

    \textbf{What You’ll Do}
    Watch 33 short video clips (about 4.7 minutes total), and write a caption for each clip.
    For each clip:
    \begin{itemize}
        \item A red dot indicates the wearer’s gaze.
        \item Write one sentence describing what the wearer is doing (from their perspective), e.g., "I pick up my phone from the table...".
        \item Each caption must be at least 30 characters long.
    \end{itemize}

    \textbf{Risks, Confidentiality, and Voluntary Participation}
    \begin{itemize}
        \item No known risks are associated with this study.
        \item Your responses are confidential, not linked to your identity, and securely stored.
        \item Participation is voluntary. You may stop at any time by closing the survey.
    \end{itemize}

    \textbf{Instructions after each video clip:}
    Then, imagine you are the wearer (the red dot shows your gaze) and write one sentence describing what you are doing.
    \begin{itemize}
        \item At least 30 characters
        \item Use first-person perspective (e.g., “I open the fridge and take out a Coke.”)
        \item Describe the main action and object clearly and specifically
    \end{itemize}
\end{tcolorbox}

\section{User Study Interfaces}
\label{app:user_study_interfaces}

We present the interfaces used in the user studies for annotating the Aria Everyday Activities (AEA) and SlideVQA datasets.

\begin{figure}[htbp]
    \centering
    \begin{minipage}{0.48\textwidth}
        \centering
        \includegraphics[height=8cm, width=\linewidth, keepaspectratio]{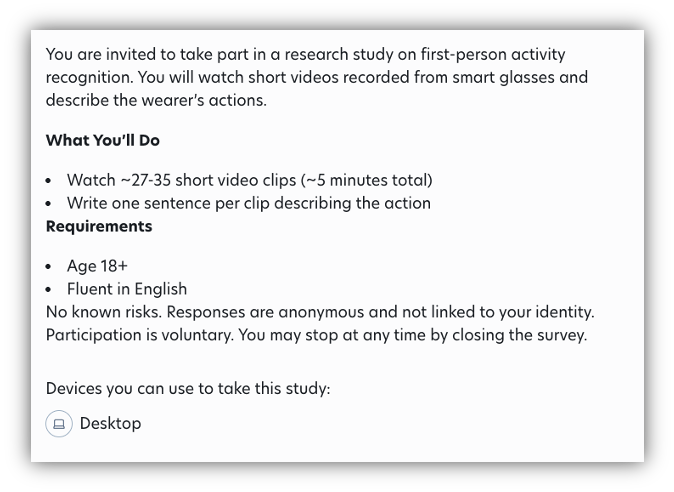}
        \caption{\textbf{Prolific landing page} for the Aria Everyday Activities (AEA) study.}
        \label{fig:aea_prolific}
    \end{minipage}
    \hfill
    \begin{minipage}{0.48\textwidth}
        \centering
        \includegraphics[height=6cm, width=\linewidth, keepaspectratio]{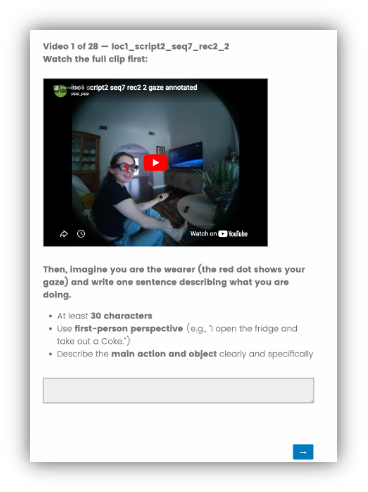}
        \caption{\textbf{Qualtrics annotation interface} for AEA, displaying a video clip with the wearer's gaze indicated by a red dot and a text box for first-person descriptions.}
        \label{fig:aea_qualtrics}
    \end{minipage}
\end{figure}

\paragraph{Aria Everyday Activities (AEA).}
The user study recruitment details page and the main annotation interface are shown in \Cref{fig:aea_prolific} and \Cref{fig:aea_qualtrics}, respectively.

\begin{figure}[htbp]
    \centering
    \includegraphics[height=8cm, width=\linewidth, keepaspectratio]{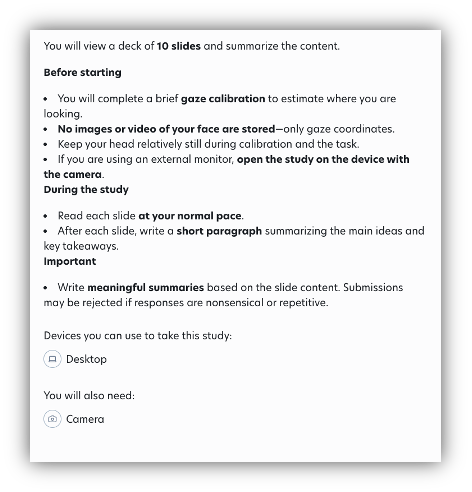}
    \caption{\textbf{Prolific study landing page} for the SlideVQA study, outlining eye-tracking calibration and response requirements.}
    \label{fig:slidevqa_prolific}
\end{figure}

\begin{figure}[htbp]
    \centering
    \begin{minipage}{0.48\textwidth}
        \centering
        \includegraphics[height=4.2cm, width=\linewidth, keepaspectratio]{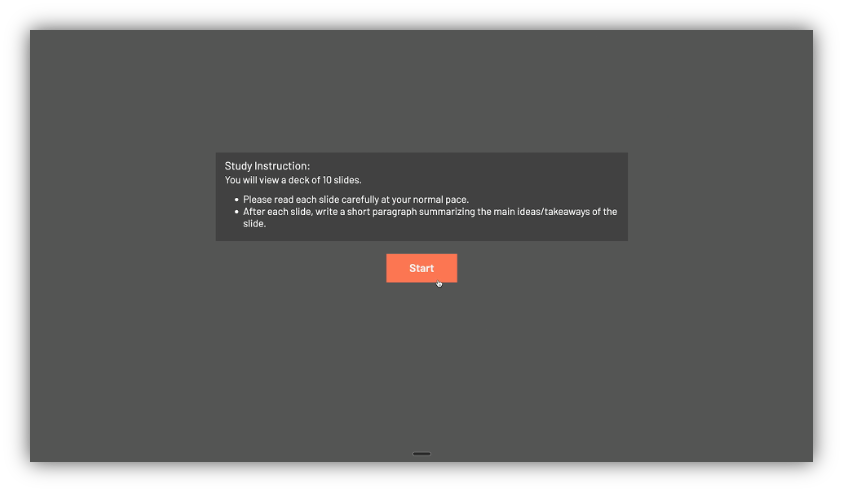}
        \caption{\textbf{RealEye study instruction screen} after eye-tracking calibration.}
        \label{fig:slidevqa_realeye_instr}
    \end{minipage}
    \hfill
    \begin{minipage}{0.48\textwidth}
        \centering
        \includegraphics[height=4.2cm, width=\linewidth, keepaspectratio]{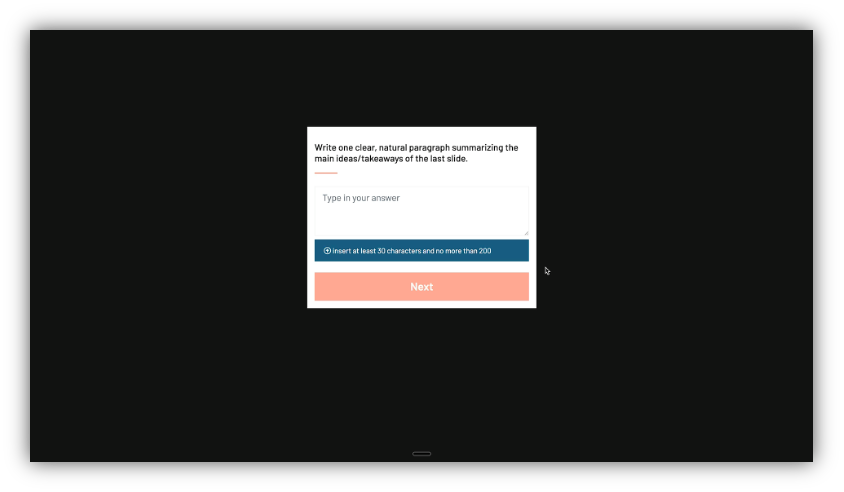}
        \caption{\textbf{RealEye annotation interface} after showing a slide to a participant.}
        \label{fig:slidevqa_realeye_task}
    \end{minipage}
\end{figure}

\paragraph{SlideVQA.}
The Prolific landing page and the RealEye eye-tracking and response collection interfaces for the SlideVQA study are shown in \Cref{fig:slidevqa_prolific,fig:slidevqa_realeye_instr,fig:slidevqa_realeye_task}.

\section{Human Annotator Demographics}
\label{app:demographics}
\paragraph{Aria Everyday Activities.}
We recruited $53$ participants via Prolific to provide activity captions for the video clips.
All participants provided informed consent prior to participation.
The cohort has a mean age of $44.3$ years ($\text{SD} = 13.1$, median $= 42$).
With respect to gender, $56.6\%$ of participants identified as female,
$37.7\%$ as male,
$3.8\%$ as non-binary or third gender,
and $1.9\%$ preferred not to say.

\paragraph{SlideVQA.}
We recruited $150$ participants via Prolific to serve as eye-tracking annotators. All participants viewed stimuli on desktop browsers using RealEye's webcam-based eye tracker. The mean eye-tracking quality grade was $4.88$ out of $6$, with $87.4\%$ of sessions rated as high quality (grade $\ge 4$). Participants and their demographic information (age, gender) were not collected as part of the anonymous Prolific study protocol. 

\newpage
\section{Prompts Used}
\label{app:prompts}
We use the following prompts for VLM candidate caption generation and VEGAS scoring.

\begin{tcolorbox}[
    enhanced,
    breakable,
    colframe=datasetAcolor!70!black,
    colback=datasetAcolor!10!white,
    colbacktitle=datasetAcolor!70!black,
    coltitle=white,
    title=\textbf{Candidate Caption Generation Prompt},
    sharp corners=south,
    boxrule=0.8mm,
    width=\textwidth,
    enlarge left by=0mm,
    enlarge right by=0mm
]
\footnotesize\ttfamily
You are a video content annotator. Generate one concise, objective paragraph describing the provided video in a natural, human-annotator style.

\medskip
Guidelines:
\begin{itemize}
    \item Describe the setting, main subjects or objects, and key actions in chronological order.
    \item Do NOT use meta phrases such as ``the video begins with,'' ``the clip shows,'' or ``the camera shows.''
    \item Start sentences directly with the observable subject or action, e.g., ``A woman in a blue dress walks...''
    \item Include only essential visual details when relevant.
\end{itemize}

Constraints:
\begin{itemize}
    \item Output only the descriptive paragraph.
    \item No interpretation, speculation, or non-observable information.
    \item Maintain a neutral, factual tone.
\end{itemize}
\end{tcolorbox}

\begin{tcolorbox}[
    enhanced,
    breakable,
    colframe=datasetCcolor!70!black,
    colback=datasetCcolor!10!white,
    colbacktitle=datasetCcolor!70!black,
    coltitle=white,
    title=\textbf{VEGAS Scoring Prompt},
    sharp corners=south,
    boxrule=0.8mm,
    width=\textwidth,
    enlarge left by=0mm,
    enlarge right by=0mm
]
\footnotesize\ttfamily
Given the video frames and the masked caption of a video: \{masked\_caption\}.
Guess all [MASK] words originally representing any words describing the video, e.g., first\_guess second\_guess. Return only the answers, without any explanation.
Do not use quotes or commas; separate tokens with a single space.
\end{tcolorbox}

\section{Caption Post-processing}
\label{app:caption_postprocessing}
Because the LLM gateway serving pipeline did not reliably strip Qwen's thinking-mode content from returned captions, we filtered Qwen outputs containing reasoning artifacts such as intermediate analysis, drafting language, or phrases like ``the user wants,'' ``identify,'' and ``let's refine.'' This reduced the Qwen SlideVQA captions from $300$ to $214$ examples at temperature $0.0$ and from $300$ to $164$ at temperature $1.0$; the remaining captions were pooled with Gemini- and GPT-generated captions for downstream SBERT analyses.

\section{Compute Resources Used}
\label{app:compute_resources}
Gemini and GPT captions were generated through their official provider APIs, while Qwen captions were generated using an on-premise deployment served through an LLM gateway.
VEGAS scores were generated using a locally deployed Gemma 4 model on a Primus inference deployment with 4× AMD Instinct MI300X GPUs (192 GB HBM per GPU), 32 CPU cores (Intel Xeon Platinum 8570), and 512 GB RAM. 
Data processing and downstream analyses were performed on a separate workstation equipped with 4× NVIDIA RTX 6000 Ada GPUs (48 GB VRAM per GPU), an Intel Xeon Gold 6346 CPU, and 251 GB RAM.

\section{Qualitative Example}
\label{app:qualitative_examples}

\begin{figure}[htbp]
    \centering
    \includegraphics[width=\linewidth, trim=0 120 0 0, clip]{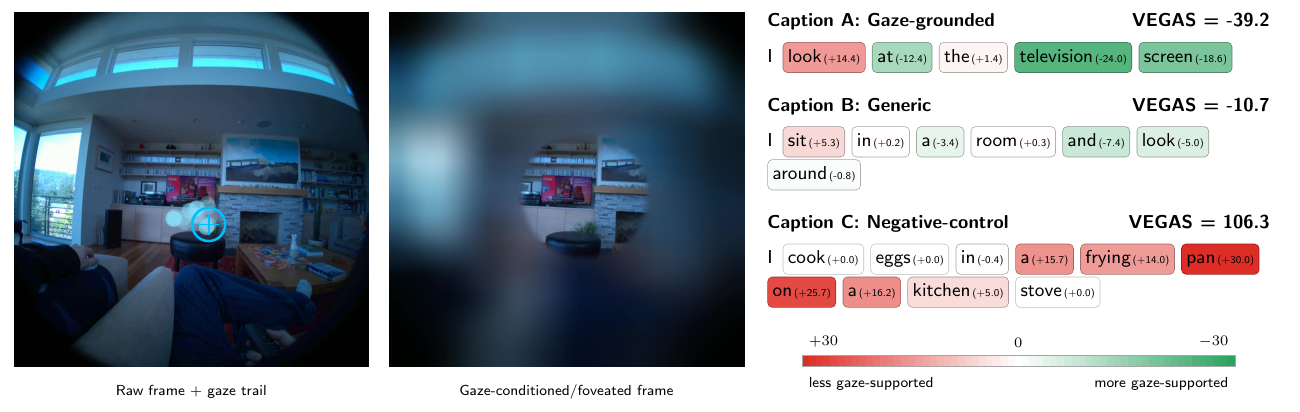}
    \caption{\textbf{Additional AEA qualitative example.} The gaze-conditioned evidence supports the caption describing looking at the television screen, while generic and negative-control captions are less gaze-supported.}
    \label{fig:aea_tv_screen_appendix}
\end{figure}

\begin{figure}[htbp]
    \centering
    \captionsetup{justification=raggedright,singlelinecheck=false}
    \begin{subfigure}{\linewidth}
        \centering
        \includegraphics[width=\linewidth, trim=0 160 0 0, clip]{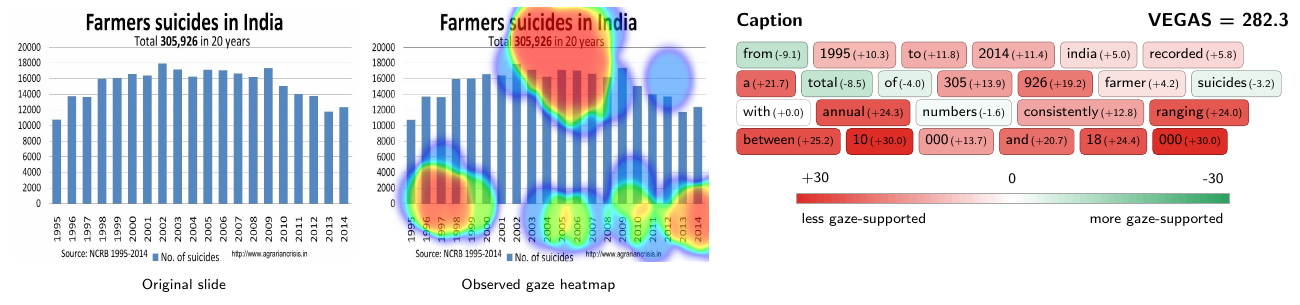}
        \caption{Chart aggregation. The caption compresses information from the full bar chart, including years, counts, and the total, while the gaze heatmap only partially covers those supporting regions.}
        \label{fig:slidevqa_bad_case_chart}
    \end{subfigure}
    \vspace{0.5em}
    \begin{subfigure}{\linewidth}
        \centering
        \includegraphics[width=\linewidth, trim=0 160 0 0, clip]{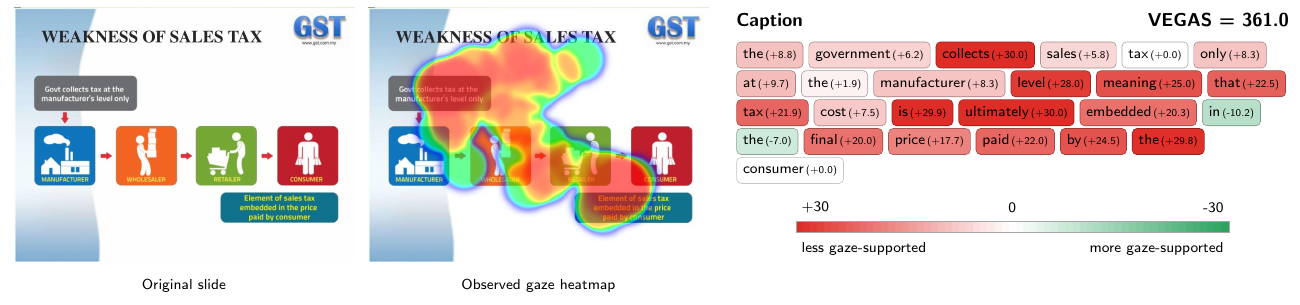}
        \caption{Causal interpretation. The caption describes the overall tax-flow mechanism, whose support is spread across multiple diagram nodes and arrows rather than a single attended region.}
        \label{fig:slidevqa_bad_case_tax_flow}
    \end{subfigure}
    \caption{\textbf{Challenging cases in SlideVQA.} Both have reasonable captions with high VEGAS scores.}
    \label{fig:slidevqa_bad_cases}
\end{figure}

This section provides three additional qualitative examples. \Cref{fig:aea_tv_screen_appendix} shows an AEA case where the gaze-conditioned frame supports a caption about the attended television screen. \Cref{fig:slidevqa_bad_cases} shows SlideVQA cases where VEGAS is less effective: the captions are reasonable, but receive high scores because they summarize global chart trends or causal relationships rather than only the locally attended regions.

\section{Ablation over top-$K$ and fallback log-probability $\lambda$}
\label{app:vegas_implementation_robustness}

VEGAS uses the scoring VLM's returned top-$K$ token log probabilities to approximate \Cref{eq:vegas_tokens}. If a target token is not in the returned set, we assign it a fallback log-probability $\lambda$. We ablate both implementation choices by selecting one caption per AEA clip from the same fixed candidate pool and measuring the selected captions with median SBERT similarity to the human-caption centroid and human-query Recall@$5$. We also report a random-candidate baseline that samples one caption uniformly from the same candidate pool, averaged over $1000$ seeds. We use \texttt{Qwen2.5-VL-32B-Instruct} as an additional scorer because it is a strong open-weight VLM of comparable scale that can be deployed locally.

\begin{table}[h]
    \centering
    \caption{\textbf{Ablation over top-$K$ and fallback log-probability $\lambda$ in AEA.} Each cell reports median SBERT / Recall@$5$ for VEGAS-selected captions under a top-$K$ and fallback log-probability setting. The random-candidate baseline samples one caption uniformly from the same candidate pool and is independent of scorer, top-$K$, and $\lambda$.}
    \label{tab:vegas_implementation_robustness}
    \begin{tabular}{llccc}
        \toprule
        & & \multicolumn{3}{c}{Fallback log-probability $\lambda$} \\
        \cmidrule(lr){3-5}
        Scorer & Top-$K$ & $-10$ & $-20$ & $-30$ \\
        \midrule
        \texttt{Gemma-4-31B-IT} & $5$  & $0.736$ / $0.509$ & $0.735$ / $0.511$ & $0.737$ / $0.517$ \\
                       & $10$ & $0.736$ / $0.514$ & $0.739$ / $0.522$ & $0.738$ / $0.523$ \\
                       & $20$ & $0.739$ / $0.515$ & $0.737$ / $0.508$ & \textbf{$0.741$ / $0.523$} \\
        \midrule
        \texttt{Qwen2.5-VL-32B-Instruct} & $5$  & $0.736$ / $0.524$ & $0.736$ / $0.519$ & $0.736$ / $0.521$ \\
                                & $10$ & $0.736$ / $0.505$ & $0.736$ / $0.508$ & $0.736$ / $0.506$ \\
                                & $20$ & $0.740$ / $0.513$ & $0.725$ / $0.511$ & \textbf{$0.725$ / $0.509$} \\
        \midrule
        \multicolumn{2}{l}{Random candidate caption, $1000$ seeds} & \multicolumn{3}{c}{$0.727$ / $0.484$} \\
        \bottomrule
    \end{tabular}
\end{table}

\Cref{tab:vegas_implementation_robustness} shows that downstream performance changes little across this grid. 
All settings are above the random-candidate baseline ($0.727$ median SBERT and $0.484$ Recall@$5$), indicating that VEGAS improves over chance selection from the candidate pool. Thus, while $K$ and $\lambda$ change the absolute VEGAS scale, the semantic and retrieval quality of the selected captions is not sensitive to these implementation details.

\end{document}